# Dependency Grammar and the Parsing of Chinese Sentences*

**LAI Bong Yeung Tom**

Department of Chinese, Translation and Linguistics
City University of Hong Kong
Department of Computer Science and Technology
Tsinghua University, Beijing
E-mail: cttomlai@cityu.edu.hk

**HUANG Changning**

Department of Computer Science and Technology
Tsinghua University, Beijing

## Abstract

Dependency Grammar has been used by linguists as the basis of the syntactic components of their grammar formalisms. It has also been used in natural langauge parsing. In China, attempts have been made to use this grammar formalism to parse Chinese sentences using corpus-based techniques. This paper reviews the properties of Dependency Grammar as embodied in four axioms for the well-formedness conditions for dependency structures. It is shown that allowing multiple governors as done by some followers of this formalism is unnecessary. The practice of augmenting Dependency Grammar with functional labels is discussed in the light of building functional structures when the sentence is parsed. This will also facilitate semantic interpretion.

## 1  Introduction

Dependency Grammar (DG) is a grammatical theory proposed by the French linguist Tesniere.[1] Its formal properties were then studied by Gaifman [2] and his results were brought to the attention of the linguistic community by Hayes.[3] Robinson [4] considered the possiblity of using this grammar within a transformation-generative framework and formulated four axioms for the well-formedness of dependency structures. Hudson [5] adopted Dependency Grammar as the basis of the syntactic component of his Word Grammar, though he allowed dependency relations outlawed by Robinson's axioms.

Dependency Grammar is concerned directly with individual words. The 'grammar' is about what companions a word can have by contracting an asymmetric head-modifier (governor-dependent) kind of relation with them. For example, using Hudson's convention of letting arrows point from heads to modifiers, we have the following Chinese sentence from [6]:

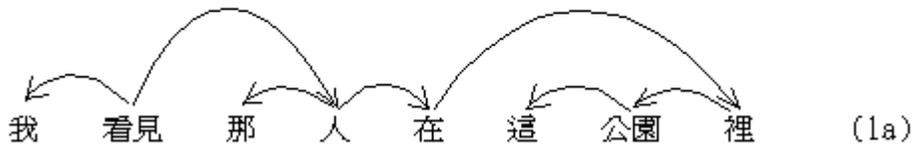

The main (or central) element, in Hayes' terminology, of the whole sentence is 看見. Its immediate dependents are 我 and 人, with 人 in turn having dependents of its own. It can be readily seen that this kind of analysis of a sentence is not unlike immediate constituency (IC). The main feature of DG that distinguishes it from IC analysis is that of the three 'top-level' elements of (1), 看見 is different from 我 and 人 in the sense that it is the 'head' element. This captures the traditional view that the two subordinate elements are 'arguments' of the main 'predicate'.

Another important observation is that when we use a phrase-structure grammar (PSG) to formalize an IC kind of analysis, we obtain for (1), for example, a phrase marker (PM), with the node labels left out, like the following from [7]:

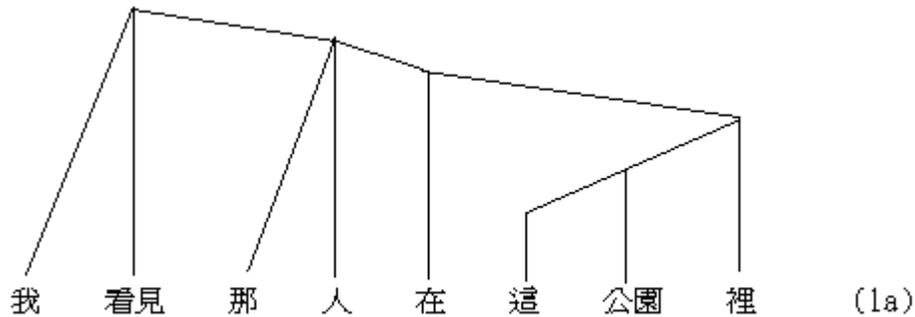

In this phrase marker, intermediate nodes are required to represent the intermediate constituents at various depths of the tree. With a DG analysis, on the other hand, the dependency structure (DS) in (1) is equivalent to the following tree representation [6]:

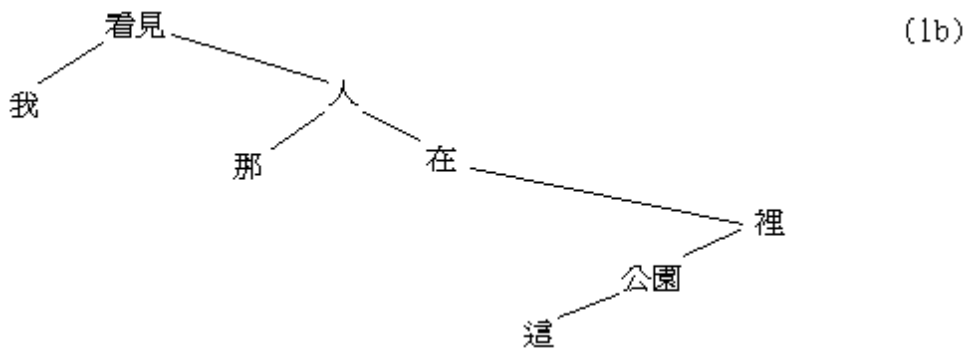

We have no unnecessary intermediate nodes in (1a). This saves space and, more importantly, provides a mechanism for the head of the sub-constituents to be identified.

The correspondence between PM's like (1a) and DG's like (1) or (1b) is many-to-one. [7] gives a procedure to derive from a PM, in which heads of sub-constituents can be identified, into a DS. They also give a procedure to derive from a DS like (1) or (1b) a 'minimal' PM like (1a).

DG has been used in linguistic analysis [8] and natural language parsing [9, 10] for languages with relatively free word order like Japanese and Korean. There have also been attempts to analyze Chinese sentences using DG [11] and corpus-based statistical parsing techniques. [12, 13, 14, 15] Efforts have been made to incorporate syntax-motivated rules in the parsing algorithm. [16, 17] Unification-based techniques can also be used in DG parsing. [18]

## 2  Formal Properties of Dependency Grammar

While exploring the possibility of using DG as the base component of a transformational-generative formalism,[19] Robinson gives four axioms for the well-formedness of dependency structures:[4]

> (a) One and only one element is independent;
> (b) all others depend directly on some element;
> (c) no elements depends directly on more than one other;
> (d) if A depends directly on B and some element C intervenes between them (in linear order of string), then C depends directly on A or on B or some other intervening element.

For ease of reference, we shall call (a) to (d) **A1** to **A4**. These four axioms are adequate for sanctioning dependency structures generated by DG's. Followers of DG in China have nevertheless introduced a fifth one:[13]

> **A5** An element cannot have dependents lying on the other side of its own governor.

This addition is formally a corollary of A4, but it makes inspection of possible dependency relations easier in many circumstances.

Robinson [4] intends to use DG to generate a 'structure free' core of natural language, so that transformations can apply to this core to yeild the entire language. She argues that to account for the generation of 'structure-sensitive' strings like 'aabaa' using a DG formalism will require the formulation of 'cumbersome' and 'ad hoc' rules. She follows Hayes [3] in using rules like the following for a DG:

> (a)   X(A,B,C,...,H,*,Y,...,Z)          (H1)
> (b)   X(*)                               (H2)
> (c)   *(X)                               (H3)

Rule (a) allows a governing 'auxilliary alphabet' X to have A, B, ..., Z as dependents. The elements A, B, ..., Z will occur in the linear order as given, with the governor situated between H and Y. Rule (b) says that the 'terminal alphabet' corresponding to X occurs without any dependents. Rule (c) says that X occurs without any governor. In other words, it is the 'main' or 'central' element of the given 'phrase'.

Gaifman [2] proves that the class of utterances that can be generated by rules like (a), (b) and (c) above is the class of context-free languages. He also proves that every DG is 'strongly equivalent' to an IC (read PS for 'phrase-structure') grammar in the sense that:

> (a)   they have the same 'terminal alphabet';
> (b)   for every string over that alphabet, every structure attributed by either grammar corresponds to a structure attributed by the other.

An IC (read 'PS') grammar with a property established by Gaifman is however not strongly equivalent to any DG.

Without going into technical details, Gaifman's results can be appreciated by noting that all DG's like (1) or (1b) in the previous section have corresponding PM's like (1a), while PM's (not being qualified by any properties) do not have corresponding DS's.

## 3   Analysis of Chinese Sentences

The complications mentioned in the previous section aside, DG as embodied in Robinson's axioms can be taken as a good attempt at capturing syntactic structure in language. The head-modifier relation is certainly captured. Besides, it should be noted that A4 captures the non-crossing constraint well-documented in PS-based grammar formalisms. To see this, let us look at the following two DG analyses for the same sentence:

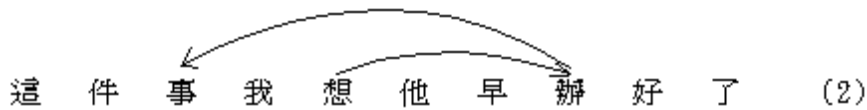

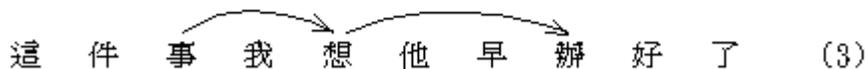

DS (3) complies with the axioms of Robinson. It corresponds to the following well-formed PM:

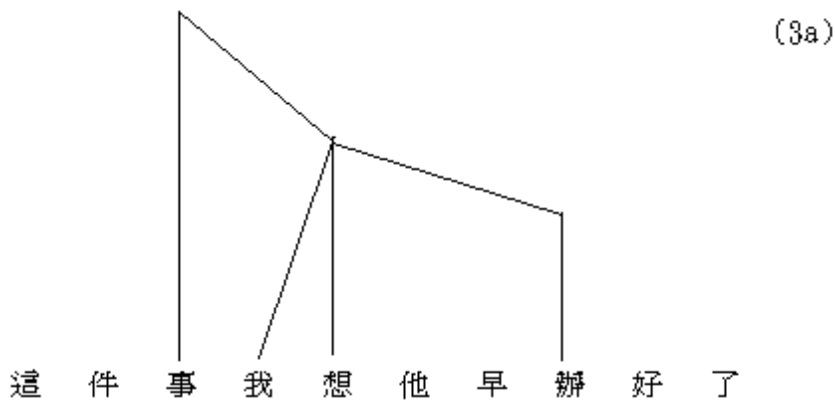

DS (2) violates the corollary A5 (and hence the axiom A4). It corresponds to an ill-formed PM with two instances of cross-over:

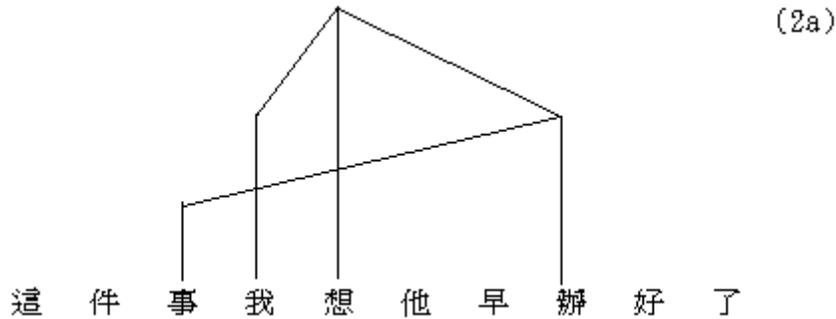

This example involves long-distance dependency. It is obviously outside the core of Robinson,[4] who will certainly require the use of transformations to account for this sentence.

Allowing (3) and (3a) and outlawing (2) and (2a) makes perfect sense when one is parsing 'surface' strings. Probing into deeper syntactic, functional or semantic relationships, it may be reasonable for 看 to be 'governed' by 看, but in the surface structure, there should be nothing wrong in letting the topic 看 be the main element of the sentence. It should be noted therefore that, like PS grammars, DG, at least as embodied in Robinson's axioms, are for surface syntactic structures.

## 4 Adding a Functional Element

Using DG to account for only surface syntactic dependencies is in no sense too restrictive. It may be noted that in Lexical-Functional Grammar (LFG),[20, 21] context-free phrase structure rules are used to account for surface syntactic structures. Deeper dependencies are taken care of by a functional element. To account for (3) and (3a), LFG uses the following functionally annotated rule:[22]

(LF1)  S'  →   XP        S
              (↑TOPIC=↓)  ↑=↓

LFG does not bother about which constituent is the syntactic head. In (LF1), S is the functional head. When the functional structure is being built up, the topic 看 is moved down to a position subordinate to the main predicate representing 看. Its functional relationship with 看, the embedded predicate, is established by means of functional control.

In fact, we claim that it is possible to add a functional element to DG. Using the set of dependency relation labels in [23], sentence (1) is actually analyzed in [6] as follows:

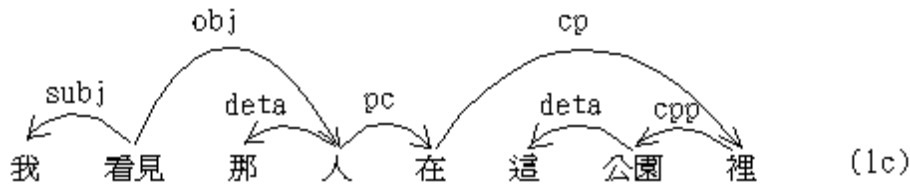

The labels of the dependency arcs can be seen to have a functional nature. With this kind of mechanism, we can also label the arcs in DS (3) to show that 看 bears a

'comment' relationship to 看 functionally. A functional control mechanism, together with lexical information that 看 needs an object, will be able to assign the governor of the 'comment' relation its proper functional role in the subordinate clause.

Another example, from [23], involving control is:

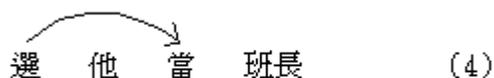
選 他 當 班長 (4)

A functional element is incorporated in the form of labels of dependency relations in all major efforts using DG to parse natural languages. When Chinese sentences are parsed in [12, 13, 14, 16, 17], the functional labels are there when the parse is produced.

The DG's of Hayes [3] and Robinson [4] are of a purer syntactic nature. But, using LFG as model again, functional annotations can be added to the Hayesian DG rules. (H1) can be annotated as follows:

(a')      X(A(fa), B(fb), ..., *, ..., Z(fz))      (H1')

where fa, fb, ..., fz are short-hands for LFG-like functional annotatins fa(X) = A, etc., reading A is X's fa.

Unlike LFG, the syntactic head and the functional head must be the same in annotated Hayesian DG rules. Functional annotations can only be attached to non-heads in Hayesian DG rules, and instead of adding annotation to the topic XP as in (LF1), we are forced to have:

(DG1)     X(*, Y(COMMENT))

But (DG1) and (LF1) differ essentially only in polarity. There should not cause any difficulty for the formalism. It can thus be concluded that DG's can be augmented with a functional element so that syntactic and functional structures are obtained at the same time when a sentence is parsed.

Availability of the functional structures will also facilitate semantic interpretation as in LFG. For (3), the functionally annotated DS:

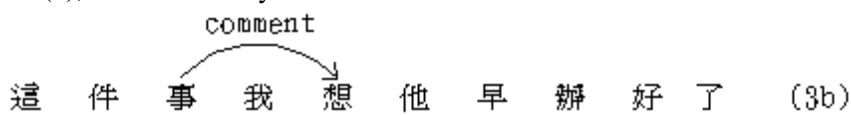
這 件 事 我 想 他 早 辦 好 了 (3b)

will yield a semantic structure like:

comment('看', '看'('看', '看'...'('看', x)))      (3c)

where X is co-referential with '看'. The exact semantic roles of the arguments of the predicate '看' can be obtained from lexical information.

## 5  Against Multiple Heads

We have been adhering to Robinson's four axioms so far. It should be noted that Hudson [5], in contravention to these axioms, allows multiple governors ('head-sharing' in his terminlogy) in the analysis of sentences like:

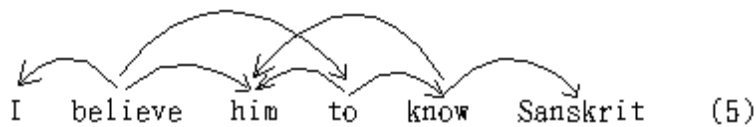

In this analysis, *him* is governed by *believe* and *know* (as well as by *to*), violating Robinson's axiom A3. This is actually not necessary because underlying semantic dependencies can be dealt with by means of functional control as we have done.

Hudson's head-sharing also gives leads to cross-dependency, in violation of axiom A4, in his analysis of the following Dutch sentence from [20]:

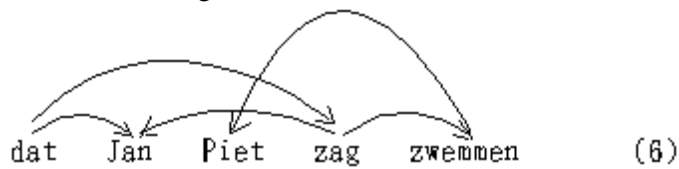

The non-crossing constraint as embodied in axiom A4 is violated as *Piet* is governed by *zwemmen* while *zag*, which lies between them, governs *Jan*. This situation will not arise if we stick to the principle that DG only takes care of surface syntactic structures and let deeper dependencies be dealt with by means of control. This observation supports our standpoint of adhering to Robinson's axioms in our interpretation of DG.

## 6  Conclusion

We have discussed how DG grammar can be augmented with LFG-style functional annotation while abiding by a set of axioms that effectively make Dependency Grammar capable of dealing with only surface syntactic dependencies. More complex functional and semantic dependencies can be dealt with by means of control mechanisms. The availability of functional structures will also facilitate semantic interpretation.

* This study is supported by the Natural Science Foundation, China.